%% file: main.tex
\documentclass[sigconf, anonymous=false]{acmart}

\usepackage{booktabs} 

\setcopyright{acmcopyright}

\usepackage{slashbox}
\acmDOI{10.1145/3555776.3577603}

\acmISBN{978-1-4503-9517-5/23/03}

\acmBooktitle{The 38th ACM/SIGAPP Symposium on Applied Computing (SAC '23), March 27-March 31, 2023, Tallinn, Estonia}
\setcopyright{licensedothergov}\acmConference[SAC '23]{The 38th ACM/SIGAPP Symposium on Applied Computing}{March 27-April 2, 2023}{Tallinn, Estonia}

\acmYear{2023}
\copyrightyear{2023}

\acmArticle{4}
\acmPrice{15.00}

\usepackage{centernot}
\usepackage{caption}
\usepackage{subcaption}
\usepackage{multirow}
\usepackage{soul}

\usepackage{adjustbox}
\usepackage{caption}    
\AtBeginDocument{%
  \providecommand\BibTeX{{%
    \normalfont B\kern-0.5em{\scshape i\kern-0.25em b}\kern-0.8em\TeX}}}

\usepackage{tikz}
 
\usepackage{xcolor}
\usepackage{MnSymbol}

\title{Dynamic Named Entity Recognition}

\author{\small{Tristan Luiggi*\dag, Laure Soulier\dag $\smallstar$, Vincent Guigue \ddag, Siwar Jendoubi*, Aurélien Baelde*}}
\affiliation{%
\institution{\small{*Upskills R\&D Choisy-le-roi, France}}
\institution{\dag Sorbonne Université, CNRS, ISIR, F-75005 Paris, France}
\institution{\ddag AgroParisTech, UMR MIA-PS }
\institution{$\smallstar$ Universit\'{e} Paris-Saclay, CNRS, LISN, Orsay, France}
}
 \email{firstname.lastname@{upmc.isir.fr;~ upskills.ai;~ agroparistech.fr}}

\begin{document}

\begin{abstract}

Named Entity Recognition (NER) is a challenging and widely studied task that involves detecting and typing entities in text. So far, NER still approaches entity typing as a task of classification into universal classes (e.g. date, person, or location). Recent advances in natural language processing focus on architectures of increasing complexity that may lead to overfitting and memorization, and thus, underuse of context. Our work targets situations where the type of entities depends on the context and cannot be solved solely by memorization. We hence introduce a new task: Dynamic Named Entity Recognition (DNER), providing a framework to better evaluate the ability of algorithms to extract entities by exploiting the context. The DNER benchmark is based on two datasets,  DNER-RotoWire and DNER-IMDb. 
 We evaluate baseline models and present experiments reflecting issues and research axes related to this novel task.\\

\end{abstract}

%
%



\maketitle



\section{Introduction} \label{section:intro}

Information extraction (IE) is a widely studied subfield of natural language processing (NLP) that aims to extract structured information from text. The task of Named Entity Recognition (NER) \cite{lafferty2001conditional,kuru-etal-2016-charner,lample-etal-2016-neural} is dedicated to recognize and classify entities (e.g. people, places, dates). NER usually constitutes the foundation of IE systems and  strongly impacts the performance of higher-level tasks such as the extraction of relations between entities or the construction of knowledge graphs \cite{DBLP:journals/corr/abs-1812-09449}.  

Recent advances in artificial intelligence address NER as a sequence modeling task in which each word of an input text is associated with a predefined class. State-of-the-art solutions are formulated through a deep learning framework using specifically designed architectures such as biLSTM-CRF \cite{lample-etal-2016-neural} and transformers \cite{akbik-etal-2018-contextual}. These models are tested on reference benchmarks \cite{tjong-kim-sang-de-meulder-2003-introduction, Weischedel2017OntoNotesA}.

Some variations of the NER task have been proposed to handle more difficult scenarios and solve specific issues: Entity Disambiguation \cite{DBLP:journals/corr/abs-2006-00575} aims at segmenting entities and linking them to a knowledge base; Entity matching \cite{DBLP:journals/corr/abs-2010-11075} aims at predicting if two detected entities refer to the same singularity.


These different tasks (including NER) share a common drawback: they all consider an entity as a universal concept, linked to a single class, even if it may appear in different surface forms and contexts. This limits the potential of the information extracted which could be useful for more elaborated downstream tasks. As an example, assuming a text relating facts of the \textit{Amazon}  company, our objective is to introduce a variability depending on the context, which might be in this case, for instance, its different roles of \textit{seller}/\textit{buyer}. 
Indeed, \textit{Amazon} is likely to \textit{sell} a product to an \textit{individual person} but \textit{buy} from another \textit{company}. Moreover, typing an entity to a single class --in the vast majority of cases-- makes memorization a competitive approach, as shown in \cite{augenstein2017generalisation,taille2020contextualized}.

To address the contextualization of entity type and limit memorization effects,  we design in this paper a specific task, called Dynamic Named Entity Recognition (DNER), in which entity labels are sample-dependent. It consists in detecting and categorizing entities whose type varies from one sample to another and addresses more abstract concepts (e.g. winning/loosing teams in a basketball match). Related tasks encompass Entity-Aspect Linking (EAL) task \cite{10.1145/3197026.3197047} in which labels are constrained by fine-grained concepts in a knowledge base, Semantic Role Labeling (SRL) \cite{li-etal-2021-syntax} in which labels correspond to roles payed by tokens in their context (\textit{verb}, \textit{subject}...) or Event Extraction (EE) \cite{8918013} in which a token describing an event (called a trigger) is extracted and linked to associated parameters (defined by the 5W1H \textit{Who, What, Whom, When, Where and How}). Our scenario is different as the class is not specifically constrained by a knowledge base (as in EAL) or need not to be associated to explicit mentions in the input text (as in SRL or EE). The fact that labels are sample dependent ensures that a DNER system will be able to take better advantage of the sentence context and should therefore be more efficient on never seen entities and more transferable to new test data.

With this in mind, we  propose two datasets, \textbf{DNER-RotoWire} and \textbf{DNER-IMDb}. The first one is derived from the academic dataset RotoWire \cite{wiseman-etal-2017-challenges}, which contains data pairs with NBA match statistics and associated summaries. Our goal is to detect players and classify them as winners or losers based on the summary. The second dataset is based on the IMDb website from which we extracted movie synopses and the associated ordered list of actors. The objective is to classify the actors according to their credit order (1,2,3,4...). For both datasets, we ensure that entities are classified differently across several samples to make the decision context-dependent. 
Our contribution  is threefold: 

$\bullet$ \textbf{DNER task formalization} (Section 3): we formalize the task of Dynamic Named Entity Recognition, including also the simplest task of Dynamic Named Entity Typing. We also detail the main associated challenges. 
    
$\bullet$    \textbf{DNER Evaluation framework} (Sections 4 and 5): we present the built datasets,  \textbf{DNER-RotoWire} and \textbf{DNER-IMDb}, and introduce a benchmark with metrics and a set of baselines.
    
$\bullet$   \textbf{Experiments} (Section 6): we conduct a series of preliminary experiments to evaluate the difficulty of the task. We outline insights reflecting the potential of the task in terms of model design. Our evaluation framework (datasets, metrics, baselines) is available at \url{https://github.com/Kawatami/DNER}. 



\section{Related Work }

Initial works in NER relied on hand-crafted features and rule-based algorithms \cite{huffman1995learning,FASTUS6}. They mainly suffered from maintenance issues, lack of flexibility and thus high adaptation cost \cite{brill1995transformation}, leading the community to explore statistical approaches \cite{toutanova2003feature}. This marked a turning point in terms of performance, especially with the introduction of the IOB scheme (later extended to IOBES), which allowed the NER task to be treated as a sequence labeling problem \cite{ramshaw-marcus-1995-text}.
Early proposals were divided into sequence modeling approaches (Hidden Markov Model) \cite{bikel-etal-1997-nymble} and classical discriminators relying on rich contextual features \cite{sekine-etal-1998-decision, asahara-matsumoto-2003-japanese}. In this context, the CRF --conditional random field-- \cite{lafferty2001conditional}, despite the cost of the Viterbi inference, received much attention \cite{mccallum-li-2003-early,liu2011recognizing}. 

The growing interest in deep learning   over the last 10 years
\cite{Collobert:2011:NLP:2078183.2078186} had led to significant advances. 
First deep-NER models  \cite{DBLP:journals/corr/StrubellVBM17,DBLP:journals/corr/HuangXY15, DBLP:journals/corr/ChiuN15, DBLP:journals/corr/ReiCP16, DBLP:journals/corr/Rei17}  exploited the semantics introduced by word representations \cite{Collobert:2011:NLP:2078183.2078186, journals/corr/abs-1301-3781}. 
\citet{DBLP:journals/corr/HuangXY15} introduced the biLSTM-CRF model which quickly became a standard architecture
with a powerful bidirectional recurrent neural network preceding a CRF layer to model label dependencies.
This backbone has been successively improved by the incorporation of representations at the level of characters \cite{lample-etal-2016-neural} and then tokens. Finally efforts have been made toward a better exploitation of additional supervision from eternal source of information \cite{DBLP:journals/corr/abs-2107-11610, Wu2020, Shah2021}.

These progresses have pushed the community to design more challenging NER tasks such as  Entity-Aspect Linking, Event Extraction \cite{10.1145/3197026.3197047} or take over Semantic Role Labeling \cite{li-etal-2021-syntax}. These tasks require to exploit the context either by establishing a link to a knowledge base or bound tokens between them with links having a special semantic. These scenarios do not encompass real-life situations where one can deduce entity roles not explicitly described in the input (the role is not part of a knowledge base nor can be represented as a link between two tokens following the predicate-argument structure) and might vary according to the context (such as \textit{winner}/\textit{loser} in basketball matches, \textit{buyer}/\textit{seller} in contracts, \textit{etiologic}/\textit{symptomatic} treatments in medical reports).

Recently,  contextualized word representation models, such as BERT \cite{DBLP:journals/corr/abs-1810-04805}, have disrupted the NER task. 
This contextualization allows disambiguation of words while allowing a very efficient fine-tuning on most NLP tasks \cite{joshi2019bert,wiedemann2019does}, and leads to better performances in the case of NER \cite{akbik-etal-2018-contextual}.
In addition, the very fine modeling of the dependencies in the self-attention layers made it possible to dispense with the costly output CRF layer \cite{taille2020contextualized}.
Those modern language models offered a lot of opportunities \cite{rogers2020primer}: their extraction abilities improved transfer learning on NER \cite{rodriguez2018transfer}, even in the more difficult setting in which the target domain was only associated with distant supervision \cite{liang2020bond}.  

Recent studies show that the complexity of those architectures enables them to encode information as a knowledge base \cite{roberts2020much}. 
Those capacities also raise new questions: regarding entities, what is the balance between memorization and extraction? From an even more general point of view, is memorization necessary -to integrate prior knowledge- or simply a phenomenon of over-fitting that must be limited? 
Historical datasets implicitly emphasize memorization capabilities by sharing an important part of the entity set between the training and the test set \cite{augenstein2017generalisation}. This phenomenon can be set aside either by designing a non-overlapping dataset \cite{derczynski-etal-2017-results} or by investigating transfer between datasets \cite{taille2020contextualized}. 
All these works around the notion of generalization in NER serve as a basis of reflection for this article.
The current performances of language models push us to test more and more ambitious problems, this is the position of this article.


\section{The DNER Task \& challenges} \label{section:taskFormulation}

Traditionally NER is formulated as sequence tagging task \cite{DBLP:journals/corr/HuangXY15,lample-etal-2016-neural}. Inspired by this formulation, we consider a supervised text $T$ describing a single event (a basketball match or a film synopsis) in which entities (and by extension all their mentions) assume a single class within it. The text itself can be decomposed as a tuple $T= (X,Y)$:\\
\indent $\bullet$ $X$ corresponds to the raw textual data, split in $N$ tokens: $X=\{x_1, \dots, x_i, \dots, x_N\}$, each token being drawn from a vocabulary $\mathcal X$.  

\indent $\bullet$  The entities are not nested, each corresponds to a consistent block of index $I = [i:i+j]$.

\indent $\bullet$ $\mathcal{V}$ which corresponds to the set of possible tags associated to entities. For our two proposed datasets, we define $\mathcal V = \{winner,loser\}$ and $\mathcal V =\{1,2,3,4\}$, respectively. Note that these labels are not necessarily explicitly associated to any token in the input text $X$. 

 \indent$\bullet$ $Y$ stands for the set of IOBES labels associated with tokens mentions within T: $\mathcal{Y}=\{y_1, \dots, y_i, \dots, y_N\}$. 

For  the  datasets proposed, we define $\mathcal{Y}=\{[B, I, E, S]-winner, [B,I,E, S]-loser, \varnothing\}$ or $\mathcal{Y}=\{[B,I,E,S]-1,[B,I,E,S]-2,[B,I,E,S]-3,[B,I,E,S]-4, \varnothing\}$). Thus, $\mathcal{Y}$ is an extension of the label set $\mathcal{V}$ dedicated to the label sequence tagging task. 

\begin{figure*}[t]
    \includegraphics[width=0.7\textwidth]{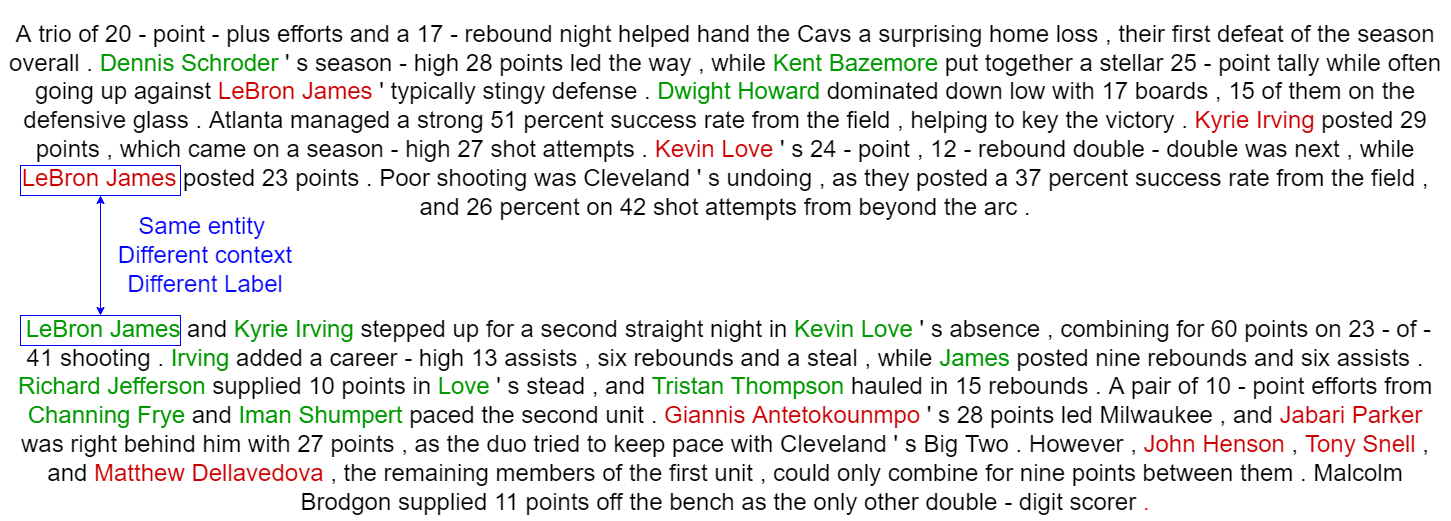}
    \caption{RotoWire preprocessed samples for the DNER task. Players highlighted in green are winners, and those in red are losers. Both samples mention the player "\textit{Lebron James}" but with different labels.}
    \label{fig:RotoWireSample}
\end{figure*}

\subsection{Tasks}
Based on these notations, we formalize  two tasks to introduce two levels of difficulty for both datasets.  We distinguish the DNET task from the DNER one, starting with the simplest. Please note that all metrics will be defined at the entity level.  

\paragraph{\textbf{Dynamic Named Entity Typing - DNET}}~\\
This task consists in classifying an already identified span of tokens indexed by $I$.
Thus, we design a function $f_{dnet}$ that makes a decision for a single entity mention within the given a context $X$,
$f_{dnet} : \mathcal{X}^N \times I \to \mathcal{V}$. 

\paragraph{\textbf{Dynamic Named Entity Recognition -  DNER}}~\\
This task corresponds to the  complete  NER, including both span identification and label assignation. The task is thus formalized as a sequence tagging: $f_{dner} : \mathcal{X}^N \to \mathcal{Y}^N$.

\subsection{Challenges} \label{sec:challenges}
Having the task formalization in mind, we can outline the different challenges of DNET and DNER:

\indent $\bullet$ \textit{Label variability}: an entity may have different labels depending on the sample, making the context influential and decreasing the informativeness of the entity's surface form. Typically, two basketball teams play each other several times, possibly with different results.
The challenge is then to focus on the language elements designating the winners and losers but not on the team membership, which would lead to overfitting. It is worth noting that while NER may also assume such variability in principle it is not constrained by design and is empirically rarely found (e.g., 97.49\% of entities in OntoNote are associated to a single label). \\
  \indent $\bullet$  \textit{Label consistency}: an entity may be mentioned several times in a text $X$, possibly in slightly varying forms. It is important to maintain label consistency per entity during the inference process.  For basketball matches, if a player is one of the winners, all his mentions must be labeled accordingly in the same sample.This challenge is also found for the NER task to a lesser extent: in the case of DNER it become crucial as the label variability takes a greater importance.\\
\indent $\bullet$  \textit{Out-of-scope entities and distant supervision}: the set $Y$ of labels does not necessarily provide supervision for all of the named entities. For example, in basketball matches, we focus only on the players while other entities are likely to appear, in particular the coaches of both teams.  This raises the question for the task and the metrics: do we need to detect this type of entity? If so, how do we label them? In this article, we focus primarily on the two aforementioned challenges and use common NER metrics. We leave the in-depth analysis of this challenge for future work.



\section{DNER Datasets} \label{section:Dataset}

In this section, we describe the construction process and bias analysis of the two introduced datasets.

\subsection{DNER-RotoWire} \label{subsection:DatasetRotowire}

\paragraph{Construction and statistics.} \label{subsection:DatasetRotowireDescritpiton}
\label{subsection:DatasetRotowireConstruction}


The Rotowire dataset \cite{wiseman-etal-2017-challenges} consists of  pairs of tabular data (match statistics) and English summaries written by sport reporters. The primary goal of this dataset is to provide a benchmark for data-to-text generation models. To fit with the DNET and DNER tasks, we reprocessed the RotoWire dataset by identifying players as entities and denoting whether they belong to the winning or the losing team (a sample is provided in Figure~\ref{fig:RotoWireSample}). The procedure is done via regexp following the distant supervision paradigm which may introduce noise in the datasets, particularly,  when names vary between tabular data and summaries; for instance, our script is designed to handle partial mentions (mentioning only the last name) but shows limitations when dealing with nicknames which would require an external knowledge base to be handled correctly.


 Finally, to allow fair comparison between models,  particularly Transformer-based approaches that are mostly limited to 512 tokens, we truncate summaries at this size limit. This implies that only 1.48 entities are removed on average per summary\footnote{Both the truncated and the full version of the datasets will be provided.}.


To measure the impact of the context memorization, we design a specific pipeline to split the dataset into train/test sets inspired by \cite{taille2020contextualized}. The goal is to separate test samples according to the increasing level of difficulty. Samples might share common properties (such as the teams involved in basketball matches) that a model could overfit and thus, bias its performance. To measure this phenomenon we divide the tests to measure performances in situations where the context (e.g., the team) have been seen and not seen during training. For this dataset we define the test set \textit{Seen} as samples in which both involved teams are seen during training, \textit{Unseen} when both teams are unseen during training, and \textit{Seen/Unseen} for which only one team is seen during training. For more details about the splitting procedure, see section~\ref{testSetSplit}.

\begin{table}[t]
    \resizebox{0.5\textwidth}{!}{
    \input{tables/roto_data_stat}}
   \caption{DNER-RotoWire  statistics. \textit{Entities} refers to the number of entity mentions, \textit{Entity tokens} to the number of tokens associated to entity mentions. Columns \textit{Winner} and \textit{loser} mention the proportion of each label category. }
   \vspace{-0.6cm}
    \label{fig:RotoWirestats}
\end{table}

\begin{table}[t]    
    \begin{minipage}{\linewidth}

\input{tables/rotoWire_Samples_stats.tex}

        \caption{Proportion of common players between sets in DNER-RotoWire. From the source (rows) that appear in the target (column).}  
        \label{fig:RotoWirestatsProp}
           \vspace{-0.6cm}
    \end{minipage}%

\end{table}

\begin{figure*}[t]
    \begin{subfigure}[t]{0.45\textwidth}
    \centering
    \includegraphics[width=\textwidth]{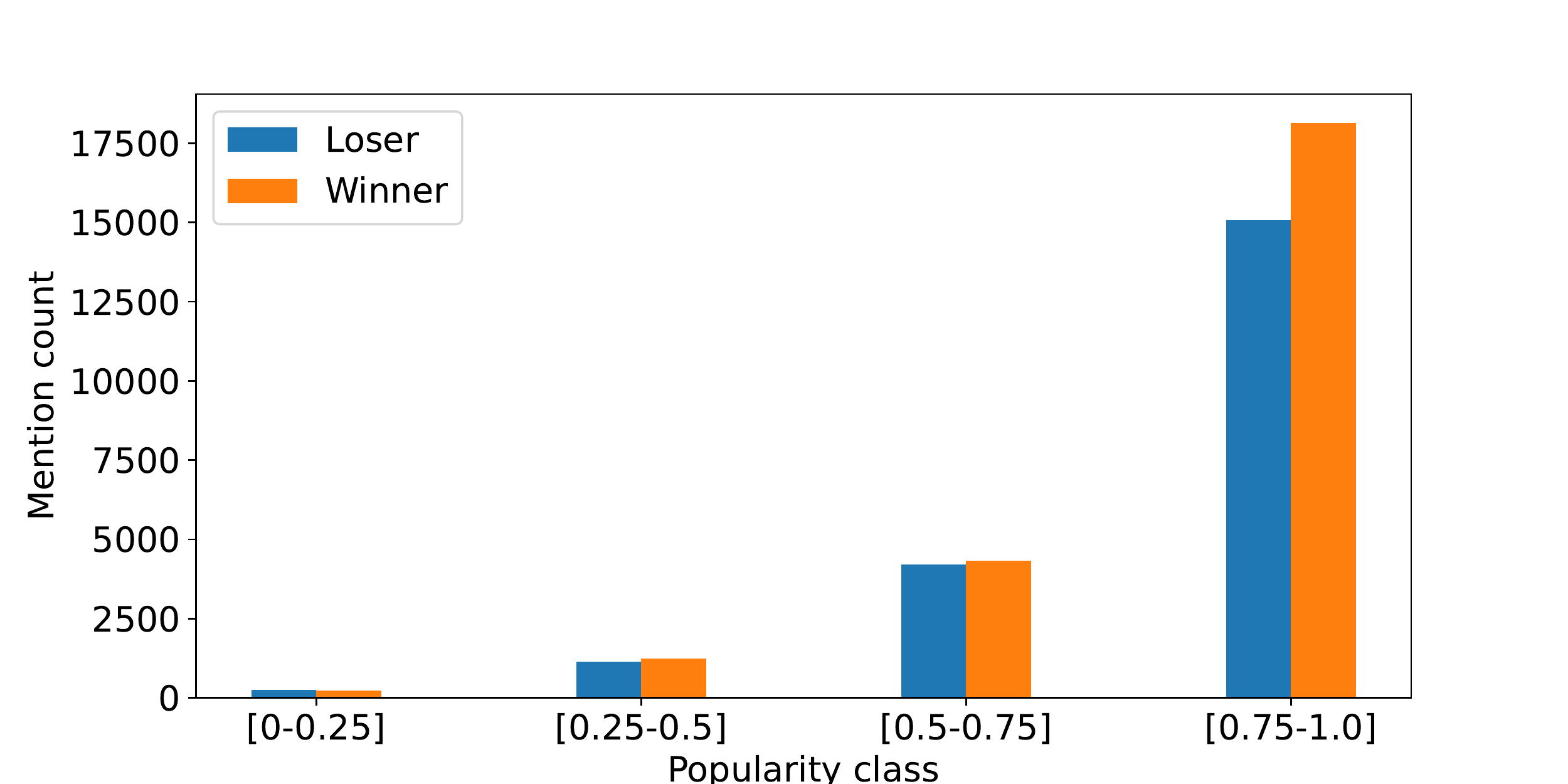}
    \caption{Winning statistics w.r.t. popularity. The quartile of most popular players tends to win more frequently.} 
    \end{subfigure}
    \hspace{0.5cm}
    \begin{subfigure}[t]{0.45\textwidth}
    \centering
    \includegraphics[width=\textwidth]{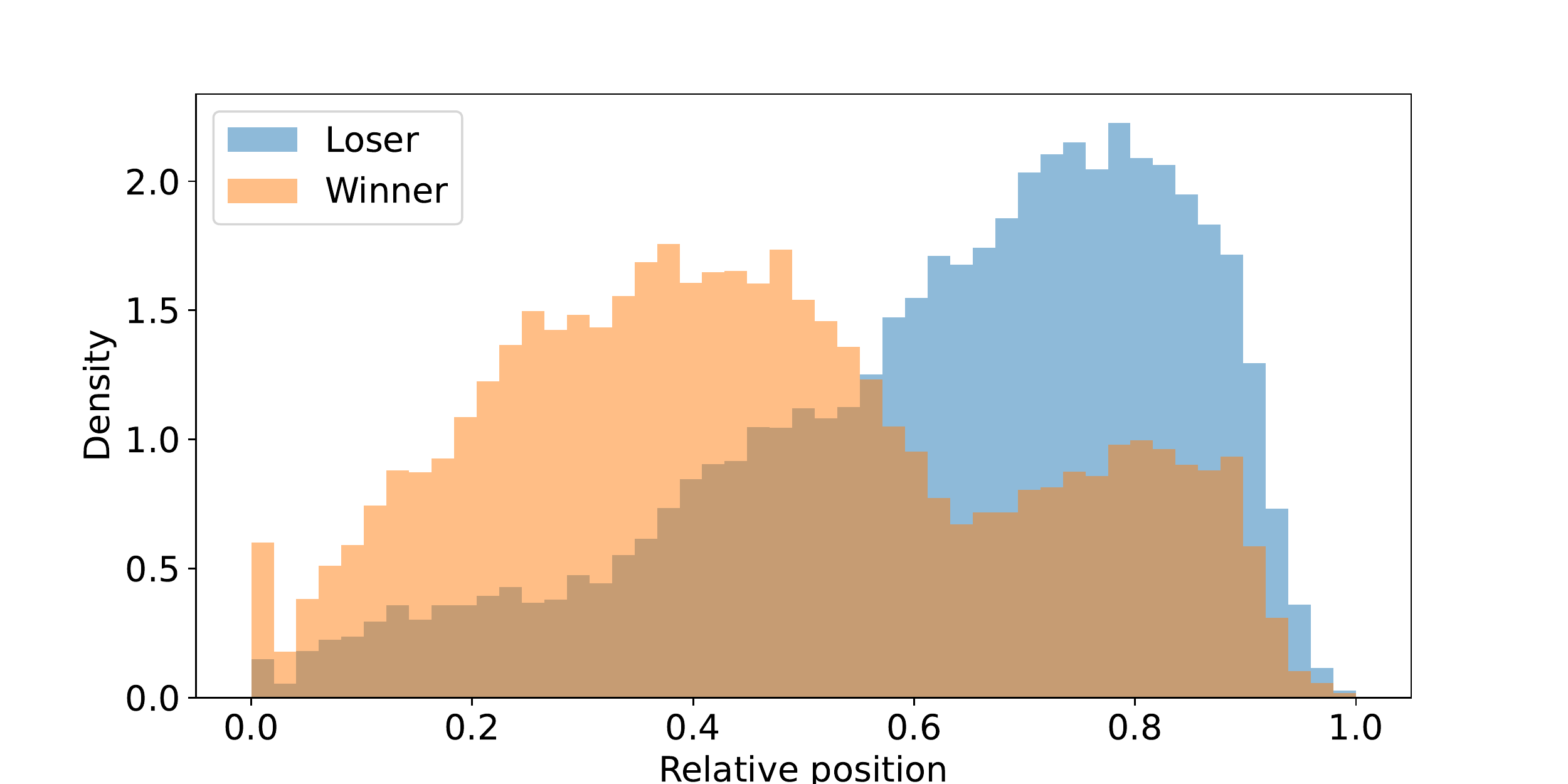}
    \caption{Player relative position distribution according to their labels. Winning players tend to be mentioned earlier.} 
    \end{subfigure}
    \caption{Analysis of popularity and relative position bias in the DNER-Rotowire dataset.}
    \vspace{-0.1cm}
    \label{fig:RotoWirestatsbias1}
\end{figure*}


Dataset statistics are provided in Table~\ref{fig:RotoWirestats}, in the resulting sets we observe an imbalance in the class distribution towards winners (55\% versus 45\%), thus better performances are expected for this class.   We checked the label variability of entities (main hypothesis of the proposed DNER task): 
we found that over 44579 total mentions, 44212 (99.17\%) belonged to players with variable labels and 367 (0.82\%) with constant ones. 
Complementary statistics about the sets are available in Table~\ref{fig:RotoWirestatsProp}.

\begin{figure*}[t]
    \centering
    \includegraphics[width=0.7\textwidth]{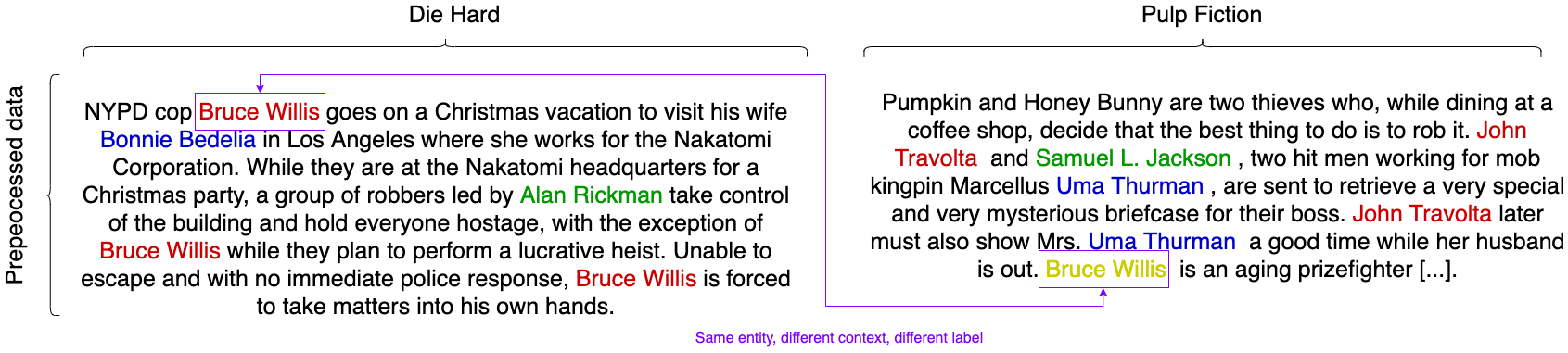}
    \caption{IMDb preprocessed samples for the DNER task. In movies  \textit{Die Hard} (left) and \textit{Pulp Fiction} (right), actors are colored w.r.t. their labels.  Red, blue, green, and yellow labels are resp. for first,  second, third and fourth actors. 
    }
    \label{fig:ImDBSample}
\end{figure*}

\paragraph{Bias analysis} \label{subsection:DatasetRotowireStatistics}
We investigate here the potential bias behind the entities regarding our DNER task. Specifically, we consider two  features:\\ 
\indent $\bullet$ \textit{Popularity}: some players are more popular than others, mainly due to their performances. This might impact the  results of matches in which they play and  lead to an over-citation ratio in  summaries. 
    
\indent $\bullet$ \textit{Position in the narrative of summaries}: 
    It seems that sport journalists tend to present the facts/players of the winning team first, then those of the losing team. 
    
In Figure~\ref{fig:RotoWirestatsbias1} (a), we provide a visual representation of the popularity bias regarding players' labels. The popularity is estimated by the ratio of a player's mentions over the total mentions in the dataset. Quartiles are then extracted to group players within four groups ranging from low to high popularity. We can observe a relative equal balancing between losing and winning mentions when players are not very popular (three first quartiles) and an unbalancing toward winning players for the most popular players. 


Figure~\ref{fig:RotoWirestatsbias1} (b) depicts the label distributions according to their relative position. Positions are normalized to range within $[0,1]$ (beginning/end of the text), 
and grouped within 50 bins. Each distribution mode  reflects the position bias: the winning players tend to be mentioned earlier on average.  



These analyses show the importance for future models to leverage the textual context to deal with label variability and avoid any bias towards popularity and position.


\subsection{DNER-IMDb} \label{section:DatasetImDB}

\begin{table*}[t]
    \input{tables/imdb_data_stat}
    \caption{
    Sample, entity and token counts for DNER-IMDb. Entity label distribution for every set.}
    \label{fig:ImDBstats}
    \vspace{-0.8cm}
\end{table*}

\begin{figure*}[t]
    \begin{subfigure}[t]{0.49\textwidth}
        \includegraphics[width=\textwidth]{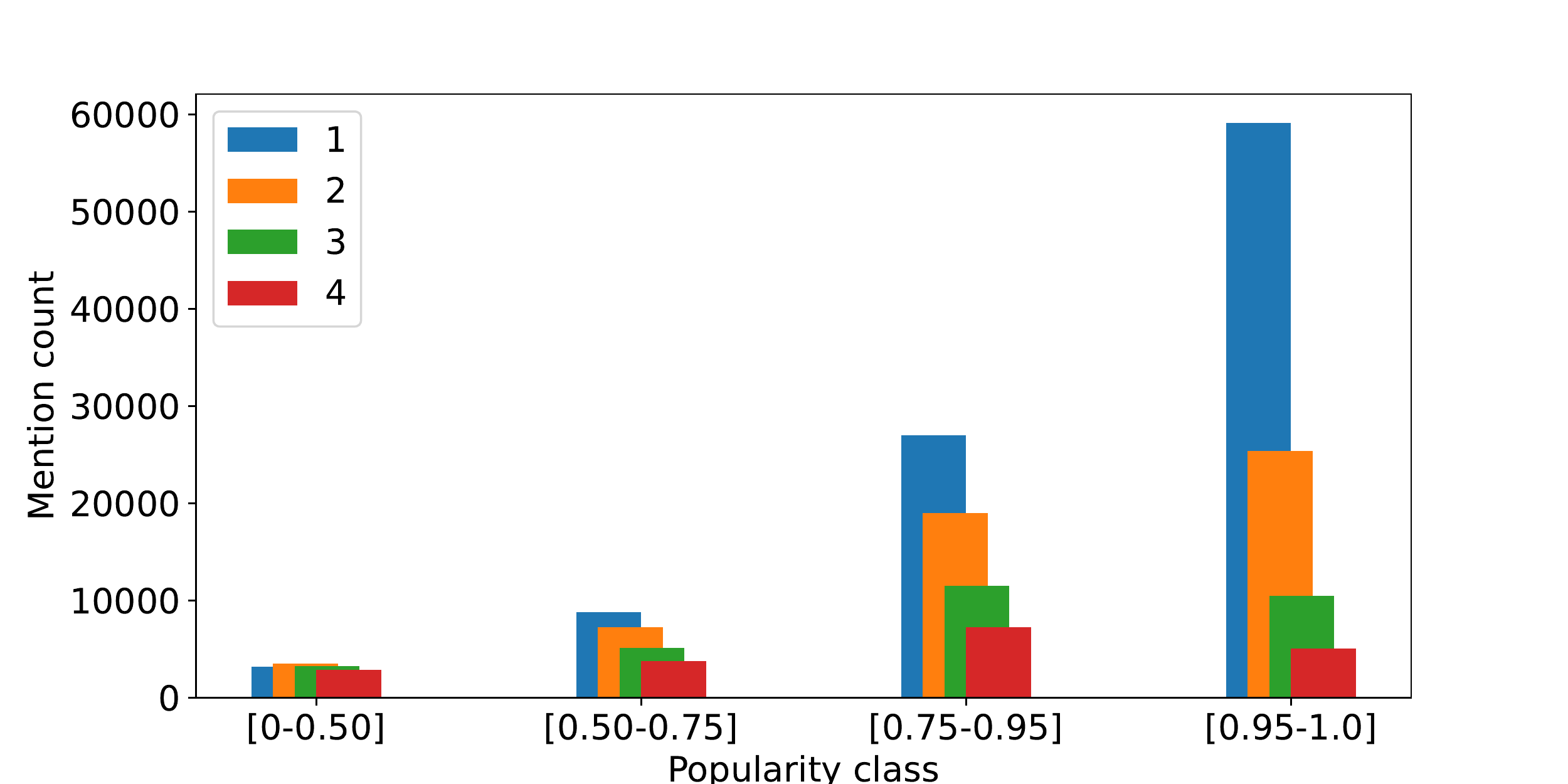}
        \caption{Credit order statistics w.r.t. popularity. Popular actors tend to take first and second place in the credit while being less expected to be in third and fourth.}
    \end{subfigure}\hfill 
    \begin{subfigure}[t]{0.49\textwidth}
        \includegraphics[width=\textwidth]{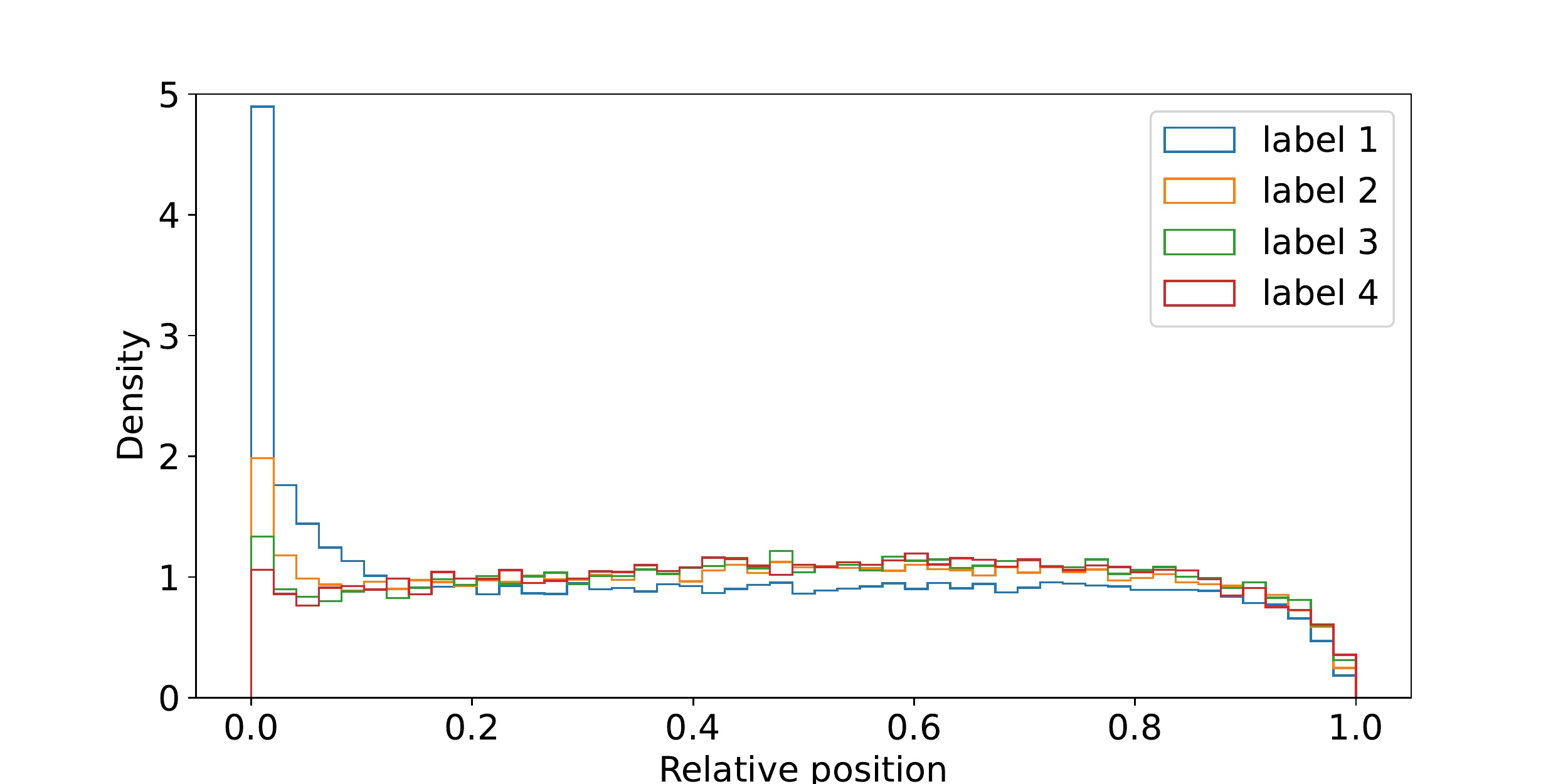}
        \caption{Actor relative position distribution according to their labels. First credited actors tend to be cited more often earlier in the summary.}
    \end{subfigure}

    \vspace{-0.4cm}
    \caption{Analysis of popularity and relative position bias in the DNER-IMDb dataset.}
    \label{fig:ImdbBias}
\end{figure*}

We  provide the DNER-IMDb dataset with the two goals: 1) adding more variability in terms of vocabulary and 2) increasing the difficulty of the task with more output classes and uncertain labels.

\paragraph{Construction and statistics} \label{subsection:DatasetImDBDescritpiton}

IMDb\footnote{\url{https://www.imdb.com/interfaces/}}\footnote{ \url{https://rapidapi.com/apidojo/api/imdb8} (as of 27/01/2022)} is an online database related to media content. We focus on movies, characterized by two main pieces of information that we use for the DNER task: 

\indent $\bullet$ \textit{Movie synopses}: these are short English descriptions of movies. 
    Synopses mention fictional characters, their importance and behavior in the films, and the relationships between the characters. A movie may have several synopses.
    
\indent $\bullet$  \textit{Character meta-data}: character information, including the actors playing those roles and their credit rank. 

To fit with our objective of dynamic labeling,  we  replace fictional characters with actor names. Indeed, an actor will appear in several movies, probably with different labels. For instance, \textit{Bruce Willis} played as first credited character in "Die Hard" but fourth in "Pulp Fiction" (see Figure~\ref{fig:ImDBSample}).  
Once the substitution is done, the designed task is to identify the credit order of all actors  given a synopsis.


In practice, we built the DNER-IMDb dataset with a raw database of 245,404 movie synopses. The labels range from the first credited actor to the eighth, but less than 1\% of the samples have more than 4 credited actors.
We only retain films with no more than 4 credited actors, produced between 1970 and 2021, and with a minimum of 600 views on the IMDb site. This ensures that synopses are written in a modern style with sufficient metadata about the actors. Moreover, actor names explicitly mentioned in the synopsis are removed (e.g. "F.B.I. trainee Clarice Starling \st{\textbf{(Jodie Foster)}} works hard [...]" - \textit{The Silence of the Lambs}). Then, we replace the names of the characters with the names of the actors using regular expressions.   There are still a few improperly formatted samples; the main errors being  (1) the mismatch between the characters' surface forms provided by the IMDb database and that found in the synopsis, resulting in inconsistent or partial permutations, and (2) mentions of characters without associated data.  Similarly as in DNER-RotoWire, we restricted the synopsis length to 512 tokens. We observed that synopses are very short on average therefore the size limit of 512 tokens has almost no impact here.

The construction of the training and test sets follows a similar procedure for the DNER-RotoWire dataset (see Section 5). As samples are only described by unique movie titles on the synopsis level (and not two teams for a match summary in the DNER-RotoWire dataset), the procedure is simplified as no \textit{seen/unseen} set is produced. 

The resulting dataset consists of 44,189 samples (a sample is provided in Figure~\ref{fig:ImDBSample}) with synopses averaging 106.89 words in length and 4.59 actor mentions. Dataset statistics are given in Table~\ref{fig:ImDBstats} and complementary statistics are available in Table~\ref{fig:ImDBstatsProp}. To assess the label variability hypothesis, we estimate 
the distribution of actors' mentions w.r.t. their number of different associated labels in the ground truth. Although there is an important number of mentions related to actors with the same label (23.9\%), we measure that most mentions are associated with entities with multiple labels (18.7\% have 2 labels, 26\% have 3 labels, and 31.4\% have 4 labels).

\begin{table}[t]

\input{tables/ImDB_Sample_stats.tex}

    \caption{Proportion of common actors between sets in DNER-IMdB. From the source (rows) that appear in the target (column).}
    \label{fig:ImDBstatsProp}
    \vspace{-1cm}
\end{table}


\paragraph{\textbf{Bias analysis}} \label{subsection:DatasetImDBStatistics}

This dataset shares common similarities with DNER-RotoWire regarding biases. We hypothesize that an actor also has a popularity factor and a position in synopses that might influence the decision process as well. Many movie synopses start by mentioning the first character (e.g. "\textbf{Mr. Cobb} a unique con artist can enter anyone's dreams [...]" - \textit{Inception}) which constitutes a potential bias. Actors' relative positions have been analyzed in Figure \ref{fig:ImdbBias} (a). We notice that first credited actors are usually mentioned earlier, specifically at the very beginning of synopses. This effect occurs for all classes, with less impact on other credit orders. Except for the beginning of synopses, we observe a homogeneous distribution of the position across labels. Similarly to DNER-RotoWire, we analyzed the popularity bias in Figure \ref{fig:ImdbBias} (b). The popularity of each actor is estimated by the ratio of her/his mentions over the number of mentions in the ground truth. We can observe that the number of mentions for classes 1 and 2 increases with the popularity level while it remains stable or decreases for classes 3 and 4. This means that popular actors are more likely to be assigned to first or second credited roles than to be assigned to last credited ones. However, when actors are not really popular, they can be uniformly assigned to any roles in the credit order. 

As for the DNER-RotoWire, this bias analysis reinforces our intuition that models need to focus on language elements to avoid learning by heart bias and not being robust to label variability.

\section{Test set construction procedure} \label{testSetSplit}

\begin{figure}[t]
        \centering
        \includegraphics[width=0.45\textwidth]{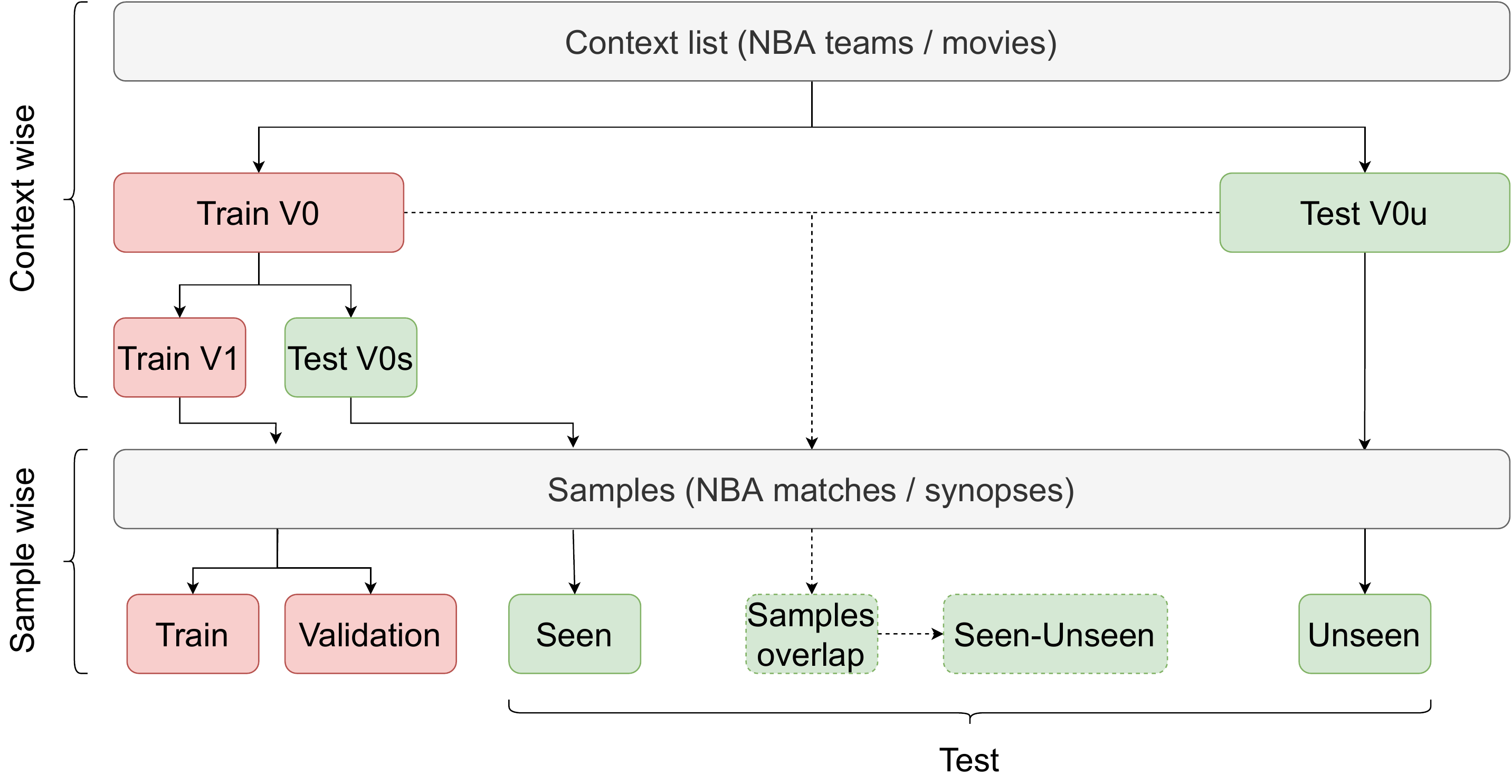}
        \vspace{-0.4cm}
        \caption{Data splitting procedure for set creation. }
    \vspace{-0.6cm}
    \label{fig:testenv}
\end{figure}

To measure the impact of context memorization, we design a specific pipeline for splitting the datasets into train/test sets. We make the hypothesis that current architecture such as LSTM or transformer are complex enough to retain entities/labels association via overfitting and not solely on context analysis. To measure this phenomenon we divide the tests to measure performances in situations where the context is known and unknown. The first set hold samples with context seen during training, we expect the best performances as this set share most of its entities with the training set (see Table \ref{fig:RotoWirestatsProp} and \ref{fig:ImDBstatsProp}) which facilitates the segmentation process, in addition, a model may overfit on datasets biases previously mentioned in section \ref{subsection:DatasetRotowireStatistics} and \ref{subsection:DatasetImDBStatistics}. The second set holds data without any common context with the training data, we expect a decrease in performances as a model cannot overfit on the context in this case in addition to be exposed to never seen entities. We, therefore, build sets to dispose of seen and unseen contexts in the test set regarding those which compose the train set. The splitting pipeline is illustrated in Figure~\ref{fig:testenv}  and includes the following main steps: 

\begin{itemize}
    \item The set of context (teams and movie titles) is split into TrainV0 and TestV0u (75\%/25\%). TestV0u is considered as the set including "unseen context".
    \item The TrainV0 is split into two new sets TrainV1 and TestV0s (75\%/25\%). TestV0s represents the set of "seen context" while TrainV1 corresponds to the training contexts.
    \item A final step is necessary to aggregate these sets at the sanple level. To do so, we process the list of samples and assign to the training fold if the current context belong to the set TrainV1 and to the test fold otherwise. 
    From the TrainV1, we split into final train/validation sets on the basis of 80\%/20\%.
    For context in the test set, we distinguish two situations : "unseen" if the context belong to the set TestV0u and "seen" if the context belong to TestV0s. Therefore, the "seen" test set is guaranteed to hold sample sharing context with ones found in the train set.
\end{itemize}

Please note that depending on the data at hand a context might be defined by one property such as unique movie title in the case of DNER-IMDb. But several properties might take place in for other data such as DNER-RotoWire.  

\section{Experiment Protocol} \label{section:Protocol}


\paragraph{Baselines} \label{subsection:XpBaseline}

We propose a set of baselines inspired by state-of-the-art NER architectures, which rely on 
Transformers supplemented either by a classical discriminator or a CRF layer for difficult cases \cite{souza2019portuguese}.

For the simpler DNET task, we design a lightweight architecture that first encodes the texts at the token level. 
Since the entities are composed of a variable number of tokens, an aggregation problem must be solved before the classification stage.
In their work, \cite{toshniwal-etal-2020-cross} study several approaches to compute such representation ranging from token selection, pooling, or attention. 
Their experiments highlight that the best is task-dependent, but the max-pooling procedure is very robust across a wide range of tasks.
Thus, we use it to compute the representations of spans associated to the entities.


For both DNET and DNER, we also consider a supplementary feature modeling the general context. By exploiting the advantages of the BERT architecture, we integrate the special \textit{CLS} token into the classifier features. We believe that this additional information can potentially be useful for our task-specific challenges such as label consistency or business knowledge modeling.

As a result, we consider 4 baselines for DNER and 2 for DNET:

%
    
%

\indent $\bullet$ \textbf{BERT-Linear}: token representations are contextualized through BERT and then classified using a linear layer.

\indent $\bullet$  \textbf{BERT-CLS}:  as BERTLinear with the additional \textit{CLS} token concatenated to each word representation before classification.
    
\indent $\bullet$  \textbf{BERT-CRF}: following the SOTA in transfer NER, we propose to add a CRF output layer to explicitly model label dependencies. This architecture is dedicated to the sequence labeling task and therefore not suited for the simple CNET task.
    
\indent $\bullet$  \textbf{BERT-CLS-CRF}:  as BERT-CRF with the additional \textit{CLS} token. 

\begin{figure}[t]
        \centering
        \includegraphics[width=0.45\textwidth]{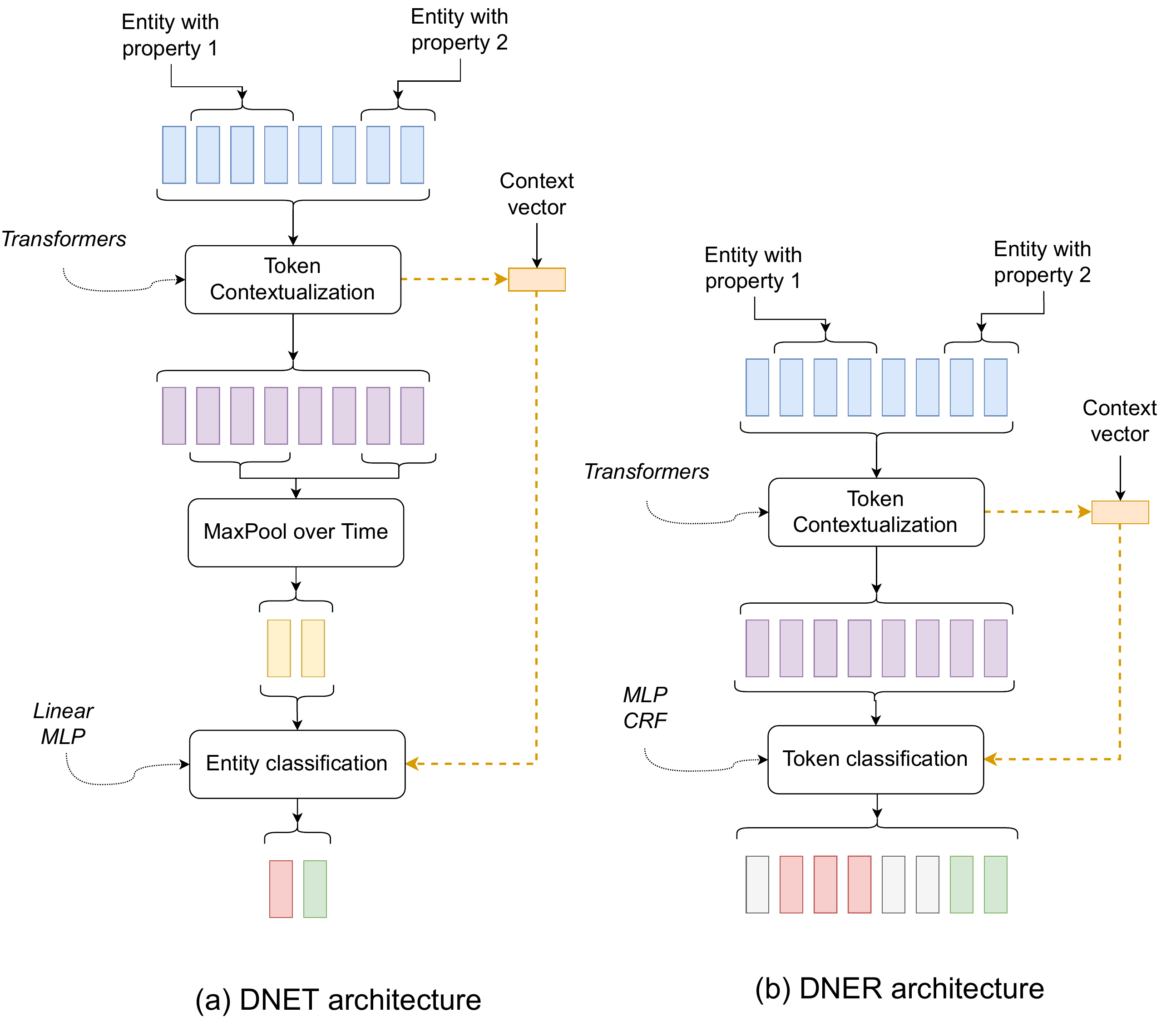}
        \vspace{-0.2cm}
        \caption{Baseline architectures} 
    \label{fig:arCHI}
    \vspace{-0.4cm}
\end{figure}

A visual representation of baseline models is provided in Figure~\ref{fig:arCHI}.



\paragraph{Metrics} \label{section:metrics}

To measure the quality of the entity classification, we make use of \textbf{$\mu F1$}. In the case of the \textit{DNER} task, the entities are extracted via the IOBES scheme in the case where it is well formatted (BEGIN token followed by INSIDE and finally END or just a SINGLE token), the entity labels are extracted via the associated class from~$\mathcal V$.

\begin{table*}[t]
\centering

\input{tables/results}
\caption{Experiment results. The $\mu F1 $ score is reported for both datasets and tasks.}
\vspace{-0.6cm}
\label{fig:results}
\end{table*}

 \begin{table*}[!t]

\input{tables/label_coherence_small.tex}

     \caption{Inconsistency analysis statistics. Entity with a single mention are ignored. S stands for the \textit{seen}  test set, S/U for the \textit{seen/unseen} test set and U for the \textit{unseen}}
     \label{tab:labelIncoherence}
     \vspace{-0.2cm}
 \end{table*}

%
To measure our ability to detect entities (whatever their classes), we  provide a span quality metric,  \textbf{Entity}. For each entity belonging to the associated reference summary, we compare each entity boundaries in the reference summary with boundaries from the \textit{begin} and \textit{end} tokens. If boundaries match, the entity has been correctly detected. We then estimate the \textit{F1} score.

To measure the entity consistency (challenges in Section \ref{section:taskFormulation}.2), we design the \textbf{inconsistency metric}. 
It compares the labels of all  mentions of the same entity within a sample. If an entity obtains the same label for all its mentions, the incoherence metric is equal to $0$. Otherwise, its value is $1$. This metric is then aggregated over all multi-mentioned entities\footnote{For clarity, all entities that are mentioned only once are removed from the calculations.} of all samples.

\section{Benchmark results} \label{section:Xp}

\paragraph{RotoWire - \textbf{DNET}}\label{subsection:XpRotoWire}

In this experiment, the goal is to classify player contextualized representations within two categories: \textit{winner} and \textit{loser}. Results are shown in Table \ref{fig:results} (left). 

We logically observe a decrease in performance when the difficulty increases from \textit{Seen} to \textit{Unseen}.
Even if the difference is limited, it is easier for a model to decide if some properties of the context have already been seen during training. 
Our two baselines perform similarly.
We can observe in Table \ref{tab:labelIncoherence} (left) the label inconsistency metric. The effect of remote supervision is visible on the ground truth with an inconsistency that varies from 2.63\% to 5.44\%. This error is amplified by the model whose inconsistencies rise to 20.54\% (\textit{unseen} test set) for BERT-Linear. This indicates that the consistency challenge is difficult to meet without explicit modeling of team membership constraints. It is interesting to note that the introduction of a general context (CLS) enables the model to significantly reduce inconsistencies (10.81\% for the \textit{unseen} test set).

\paragraph{RotoWire - \textbf{DNER}}



All DNER results are shown in table~\ref{fig:results} (right). The first conclusion from this table is that the CLS token provides a performance gain compared to the baseline \textit{Bert-Linear} architecture. This is consistent with the experiments with \textit{DNET}, where the CLS token exhibited better robustness to inconsistency. This effect could be of greater magnitude due to the IOBES scheme, which requires the classification of a larger number of classes. \textit{BERT-CRF} performs better in entity recognition, which is easily explained by the CRF layer effectively maintaining label coherence. This suggests that the CRF helps in maintaining such a factor, but is not able to correctly analyze a context. The combination of the CLS token and the CRF layer generally performs well, but the added complexity could trigger an overfitting effect. This confirms our suspicions about the added difficulty of our proposed task.

The best baseline (\textit{BERT-CLS}) proposes an average F1 performance of 0.67, mainly due to errors in typing  and precision issues; although interesting, it is clear that the many challenges mentioned in Section \ref{section:taskFormulation}.2 must be addressed in a specific way to cope with the difficulty of the DNER task.


\paragraph{IMDb - \textbf{DNET}} \label{subsection:ImDB}

On these data, the scores are globally worse than on RotoWire (Table \ref{fig:results} (right)). This is easily explained by a change from 2 to 4 classes.
The loss of performance by going from seen to unseen is much more marked than on RotoWire with more than 22 points of $\mu F1$ on average: the memorization effect brings information for all the entities with a single class.

In terms of labeling inconsistencies (Table \ref{tab:labelIncoherence} (right)), the problem is virtually absent from the ground truth. On the test data, the inconsistencies rise to 8.64\%. While this figure again shows the need to specifically address this issue, it is still far below that of RotoWire. This difference is easily explained: the number of entities per document is much lower on IMDb (4 against 9 on average on unseen) and this intrinsically reduces the risk of inconsistencies.


\paragraph{IMDb - \textbf{DNER}}

Entity detection is better on IMDb than on RotoWire, but it relies heavily on the CRF layer. 
The overall improvement is probably due to the artificial aspect of the dataset, where entities always have the same surface forms and to the overlap rate between learning and testing (54.84\%, even on unseen movies).

We then return to the conclusion of the previous section: once detected, entities are hard to categorize.
The difference between seen and unseen films is very large (6 $\mu F1$ points on average) and the overall performance tops out at 0.60 of $\mu F1$.

\section{Conclusion}

This paper introduces the Dynamic Named Entity Recognition (DNER) task which aims at detecting entities and classifying them in a frame where labels are dynamic. This task raises several challenges such as label variability, label consistency and taking into account entity position or popularity bias. We provide benchmarks in the form of two supervised datasets associated with test sets of increasing difficulty. These benchmarks are provided with metrics and reference models to ensure reproducibility and to encourage the emergence of new models to address the specific challenges of the task.
Indeed, despite a reference architecture based on transformers, our analyses show that the DNER task is particularly difficult and the results obtained can be improved. The presented datasets were designed for experimental purposes  and might not be relevant for real world applications. 

\bibliographystyle{ACM-Reference-Format}
\bibliography{sample-bibliography} 

\newpage
\appendix

\end{document}

%% file: tables/roto_data_stat.tex
    \centering
    \footnotesize
    \begin{tabular}{l@{}c@{}c@{}c@{}c@{}c@{}}
        \hline
        \textbf{\footnotesize{Set}} & \textbf{\footnotesize{Samples~}} & \textbf{\footnotesize{Entities~}} & \textbf{\footnotesize{Entity tokens~}} & \textbf{\footnotesize{Winner~}} & \textbf{\footnotesize{Loser~}}  \\
        \hline
            Train               & 1532 & 14202 & 24940 & 0.54 & 0.46 \\
            Validation          & 511 & 4615 & 8086  &  0.53 & 0.47 \\
            Seen (test)         & 511 & 4748 & 8360 & 0.54 & 0.46 \\
            Seen/Unseen (test)  & 1996 & 18293 & 31776 & 0.53 & 0.47 \\
            Unseen (test)       & 303 & 2721 & 4674 & 0.54 & 0.46 \\
        \hline
    \end{tabular}

%% file: tables/rotoWire_Samples_stats.tex
    \centering
    \footnotesize
    \begin{tabular}{cccccc}
        \hline
        \backslashbox{Source}{Target} & \textbf{Train} & \textbf{Validation} & \textbf{Seen} & \textbf{Seen/Unseen} & \textbf{Unseen}  \\
        \hline
            \textbf{Train}               & 100    & 96.71 & 97.05 & 98.82      & 20.484  \\
            \textbf{Validation}          & 99.28 & 100    & 97.50 & 98.84      & 19.60  \\
            \textbf{Seen (test)}         & 99.22 & 96.80 & 100    & 98.79      & 21.89    \\
            \textbf{Seen/Unseen (test)}  & 70.72 & 67.97 & 66.68 & 100         & 59.43    \\
            \textbf{Unseen (test)}       & 40.94 & 37.11 & 35.51 & 99.52      & 100     \\
        \hline
    \end{tabular}

%% file: tables/imdb_data_stat.tex
\centering
    \footnotesize
        \begin{tabular}{lccccccc}

        \hline
        \textbf{Set} & \textbf{\# samples~} & \textbf{\# Entity~} & \textbf{\# Tokens~} & \textbf{credit 1~} & \textbf{credit 2~} & \textbf{credit 3~} & \textbf{credit 4}  \\
        \hline
            Train               & 13328 & 90726 & 185850 & 39.50\% & 24.76\% &  21.97\% & 13.75\%\\
            Validation          & 1725 & 12008 & 24557 &  37.40\% & 24.14\% & 23.55\% & 14.89\% \\
            Seen (test)         & 805 & 5450 & 11176 & 39.68\% & 23.92\% & 23.43\% & 12.95\% \\
            Unseen (test)       & 4030 & 27346 & 56038 & 39.38\% & 24.11\% & 22.60\% & 13.89\% \\
        \hline
    \end{tabular}

%% file: tables/ImDB_Sample_stats.tex
    \centering
    \begin{tabular}{ccccc}
        \hline
        \backslashbox{Source}{Target} & \textbf{Train} & \textbf{Validation} & \textbf{Seen} & \textbf{Unseen}  \\
        \hline
            \textbf{Train}               & 100    & 19.13 & 20.76 & 20.76  \\
            \textbf{Validation}          & 95.76 & 100    & 30.57 & 43.83  \\
            \textbf{Seen (test)}         & 97.47 & 52.37 & 100    & 50.34 \\
            \textbf{Unseen (test)}       & 54.84 & 23.12 & 15.50 & 100    \\
        \hline
    \end{tabular}
    

%% file: tables/results.tex
    \footnotesize
\begin{tabular}{|c|c||c|c|c|c|c|c|}
\hline
\multirow{3}{*}{\textbf{Models}}& \multirow{3}{*}{\textbf{Set}} & \multicolumn{3}{c|}{RotoWire} & \multicolumn{3}{c|}{IMDb}                   \\
                                &                               & DNET        & DNER & Entity   & DNET           & DNER  & Entity                  \\

\hline
\multirow{4}{*}{BERT-Linear}    & Seen                          &  0.81       & 0.66            & 0.86            &  0.67            & 0.36             &  0.58\\
                                & Seen/Unseen                   &  0.81       & 0.65            & 0.85            &  -               & -                &  -    \\
                                & Unseen                        &  0.80       & 0.63            & 0.81            &  0.45            & 0.31             &  0.56\\
\hline
\multirow{4}{*}{BERT-CLS}       & Seen                          &  0.81       & \textbf{0.67}     & 0.88            &  \textbf{0.69} &  0.37            &  0.60 \\
                                & Seen/Unseen                   &  0.81       & \textbf{0.68}   & 0.87            &  -               & -                 &  -   \\
                                & Unseen                        &  0.80       & \textbf{0.67}      & 0.85           &  \textbf{0.46}  &  0.32            &  0.58\\
\hline
\multirow{4}{*}{BERT-CRF}       & Seen                          & -           & 0.67            & \textbf{0.90}   &  -              &  \textbf{0.60}  &  \textbf{0.94}   \\
                                & Seen/Unseen                   & -           & 0.67            & \textbf{0.88}   &  -               & -                &  -   \\
                                & Unseen                        & -           & 0.66            & \textbf{0.87}   &  -               &  \textbf{0.52}  & \textbf{0.92}   \\
\hline
\multirow{4}{*}{BERT-CLS-CRF}   & Seen                          & -           & 0.61             & 0.82           &  -               &  0.56             &  0.90      \\
                                & Seen/Unseen                   & -           & 0.61            & 0.81            &  -               & -                 &  -       \\
                                & Unseen                        & -           & 0.60            & 0.79            &  -               &  0.48             &  0.88       \\
\hline

\hline
\end{tabular}



%% file: tables/label_coherence_small.tex


    \footnotesize
    \centering
        \begin{tabular}{lc|cccc||cccccc}
                        &             &  \multicolumn{4}{c||}{RotoWire} &   \multicolumn{6}{|c}{IMDb} \\
        \hline
        \textbf{Model} & \textbf{Set} & \textbf{GT} &\textbf{All} & \textbf{W} & \textbf{L}  & \textbf{GT} & \textbf{All} & \textbf{1} & \textbf{2} & \textbf{3} & \textbf{4}  \\
        \hline
         \multirow{4}{*}{Bert-Linear} & S   & 5.44\%  & 17.69\%     & 14.61\%    & 21.87\%  &  0\%    & 7.24\% & 4.47\% & 6.27\%  & 11.11\% & 9.92\%\\
                                      & S/U & 4.64\%  & 20.88\%     & 17.52\%    & 26.15\%  & - & - & - & - & - & - \\
                                      & U   & 2.63 \% & 20.54\%     & 17.96\%    & 26.31\%  &  0.31\% & 8.64\% & 5.59\% & 10.05\% & 12.02\% & 9.20\% \\
        \hline
        \multirow{4}{*}{Bert-CLS}     & S      & 5.44\% & 13.27\% & 9.23\%  & 18.75\%  &  0\%       & 5.01\%   & 3.68\% & 3.76\% & 7.20\% & 7.14\%  \\
                                      & S/U  & 4.64\% & 18.27\% & 13.40\% & 25.92\%  & - & - & - & - & - & - \\
                                      & U      & 2.63\% & 10.81\% & 7.81\%  & 17.54\%  &  0.31\%   & 5.68\%   & 4.58\% & 6.41\% & 6.43\% & 6.25\%   \\
        \hline

        \end{tabular}
